\documentclass[conference]{IEEEtran}

\IEEEoverridecommandlockouts                              % This command is only needed if 
\usepackage{graphicx} % for pdf, bitmapped graphics files
\usepackage{epsfig} % for postscript graphics files
\usepackage{mathptmx} % assumes new font selection scheme installed
\usepackage{times} % assumes new font selection scheme installed
\usepackage{amsmath} % assumes amsmath package installed
\usepackage{amssymb}  % assumes amsmath package installed
\usepackage{gensymb}
\usepackage{booktabs} % For better-looking tables
\usepackage{multirow}
\usepackage{xcolor}
\usepackage{import}
\usepackage{subcaption}
\usepackage{svg}
\usepackage{tikz}
\usepackage[letterpaper, left=0.639in, right=0.639in, bottom=1.05in, top=0.75in]{geometry}
\usepackage{eso-pic} % for absolute positioning
\def\BibTeX{{\rm B\kern-.05em{\sc i\kern-.025em b}\kern-.08em
    T\kern-.1667em\lower.7ex\hbox{E}\kern-.125emX}}
\newcommand{\etal}{\textit{et al.}}
\title{\LARGE \bf FlowCalib: LiDAR-to-Vehicle Miscalibration Detection using Scene Flows}

\author{Ilir Tahiraj$^{1,*}$, 	Peter Wittal$^{2}$, Markus Lienkamp$^{1}$%
\thanks{$^{*}$ Corresponding author {\tt\small ilir.tahiraj@tum.de}}
\thanks{$^{1}$Authors are with the TUM School of Engineering and Design, Chair of Automotive Technology, Technical University of Munich.}%
\thanks{$^{2}$ is with the TUM School of Computation, Information and Technology, Technical University of Munich.}      
        }

\begin{document}

\thispagestyle{empty}
\pagestyle{empty}

\maketitle

%\begin{figure*}[!h] % Force placement at top
%   \centering
%    \includegraphics[width=1\textwidth]{figures/Ours.png}
%    \caption{Should be done via incscape}
%    \label{fig:enter-label}
%\end{figure*}

%%%%%%%%%%%%%%%%%%%%%%%%%%%%%%%%%%%%%%%%%%%%%%%%%%%%%%%%%%%%%%%%%%%%%%%%%%%%%%%%
\begin{abstract}
Accurate sensor-to-vehicle calibration is essential for safe autonomous driving. Angular misalignments of LiDAR sensors can lead to safety-critical issues during autonomous operation. However, current methods primarily focus on correcting sensor-to-sensor errors without considering the miscalibration of individual sensors that cause these errors in the first place. We introduce FlowCalib, the first framework that detects LiDAR-to-vehicle miscalibration using motion cues from the scene flow of static objects. Our approach leverages the systematic bias induced by rotational misalignment in the flow field generated from sequential 3D point clouds, eliminating the need for additional sensors. The architecture integrates a neural scene flow prior for flow estimation and incorporates a dual-branch detection network that fuses learned global flow features with handcrafted geometric descriptors. These combined representations allow the system to perform two complementary binary classification tasks: a global binary decision indicating whether misalignment is present and separate, axis-specific binary decisions indicating whether each rotational axis is misaligned. Experiments on the nuScenes dataset demonstrate FlowCalib’s ability to robustly detect miscalibration, establishing a benchmark for sensor-to-vehicle miscalibration detection.
\end{abstract}

\section{INTRODUCTION}

Extrinsic sensor calibration is key to enabling safe object detection. It ensures an accurate and reliable spatial representation of fused data. There are two methods of extrinsic calibration for multi-sensor setups: sensor-to-sensor (S2S) and sensor-to-vehicle (S2V). S2S calibration primarily contributes to object detection and sensor fusion performance, while S2V calibration contributes to overall safety and driving tasks. S2V calibration ensures accurate object localization and environment modeling, directly influencing downstream tasks such as autonomous vehicle (AV) planning. 

\begin{figure}[t]
    \centering
    \includegraphics[width=1\linewidth]{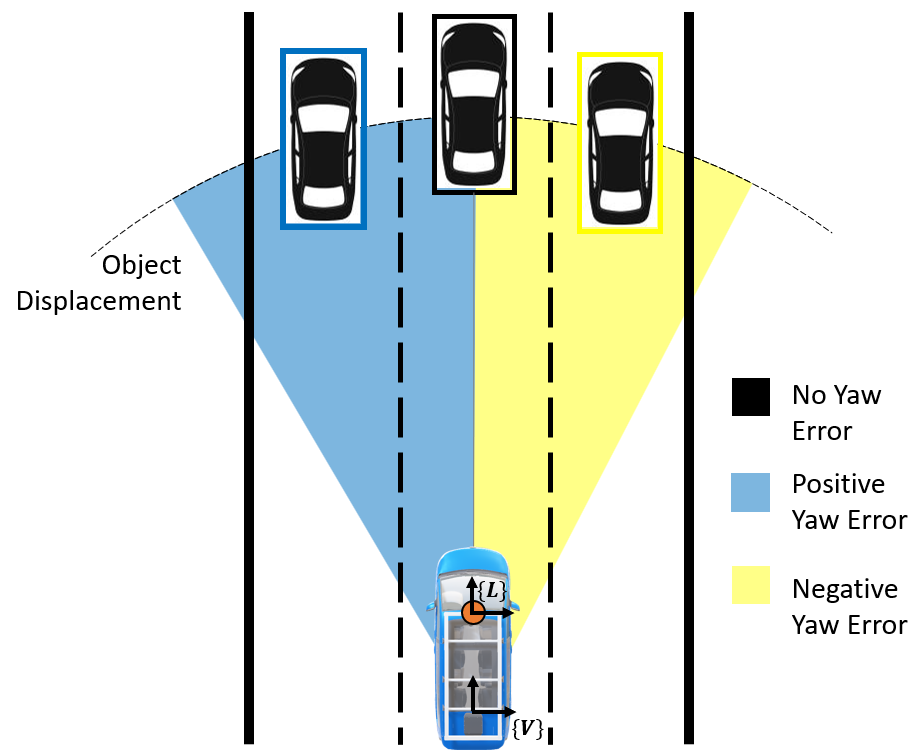}
    \caption{This is a qualitative illustration of objects in the environment model in the presence of sensor-to-vehicle yaw angle errors. Objects appearing in neighboring lanes can pose safety issues during autonomous vehicle operation. $\{V\}$ represents the vehicle frame and $\{L\}$ the LiDAR frame.}
    \label{fig:intro}
    \vspace{-0.5cm}
\end{figure}

To illustrate this, consider the following practical example of S2V yaw errors: An autonomous vehicle is driving at high speed on a highway and approaching another vehicle around 100 meters away. There are three cases of calibration error: positive yaw angle error, no yaw angle error, and negative yaw angle error. These are depicted in Fig.~\ref{fig:intro}. These errors can result in objects being detected in the lane next to the vehicle, raising safety concerns. Similarly, a vehicle driving next to the ego vehicle could be detected in the ego vehicle's lane. This can lead to comfort issues, such as unnecessary braking, as well as safety issues for vehicles driving behind the ego vehicle.

This makes S2V errors safety-critical during autonomous vehicle operation. Note that these problems can occur even with perfect S2S calibration. Although many online calibration algorithms exist that align LiDAR-camera data and meet such accuracy requirements, there is an inherent problem with online S2S calibration that must be considered for autonomous driving in the real world: detecting individual sensor miscalibration. 

Online recalibration algorithms across the field train their frameworks to regress the transformation matrix between the LiDAR and camera sensors. Training data is generated by injecting errors into the matrix that transforms LiDAR point clouds onto the image plane, thereby learning a unidirectional correction from LiDAR to the camera sensor~\cite{survey}. This approach is valid when the misalignment between the two sensors originates from a miscalibrated LiDAR sensor. However, always correcting the LiDAR points, even when the camera is miscalibrated, can worsen the situation in the real world. This problem can be solved by detecting miscalibration in one of the sensors to enable bidirectional correction.

Therefore, we propose FlowCalib, a framework for detecting LiDAR-to-vehicle miscalibration that uses object motion patterns in the presence of angular calibration errors. Specifically, we use 3D point cloud flow fields of static objects to detect extrinsic LiDAR miscalibration. Our contributions are as follows:
\begin{itemize}
    \item FlowCalib is the first LiDAR-to-vehicle miscalibration detection framework.
    \item We demonstrate that our framework can reliably detect angular calibration errors and establish a baseline for miscalibration detection on the nuScenes dataset.
    \item The detection works without the use of additional sensors, such as IMU, GPS, or camera.
    \item FlowCalib provides both global miscalibration detection and axis-specific detection, providing information on roll, pitch, and yaw angle miscalibration.
    \item The FlowCalib code will be made available as open source.
\end{itemize}
\section{RELATED WORK}
This section summarizes the work on S2V calibration and miscalibration detection. First, a sensitivity analysis of calibration errors on state-of-the-art object detectors is presented to motivate the detection of LiDAR miscalibration. Next, we provide an overview of current miscalibration detection and sensor-to-vehicle calibration algorithms in general.

\begin{figure}[h]
    \centering
    \includegraphics[width=1\linewidth]{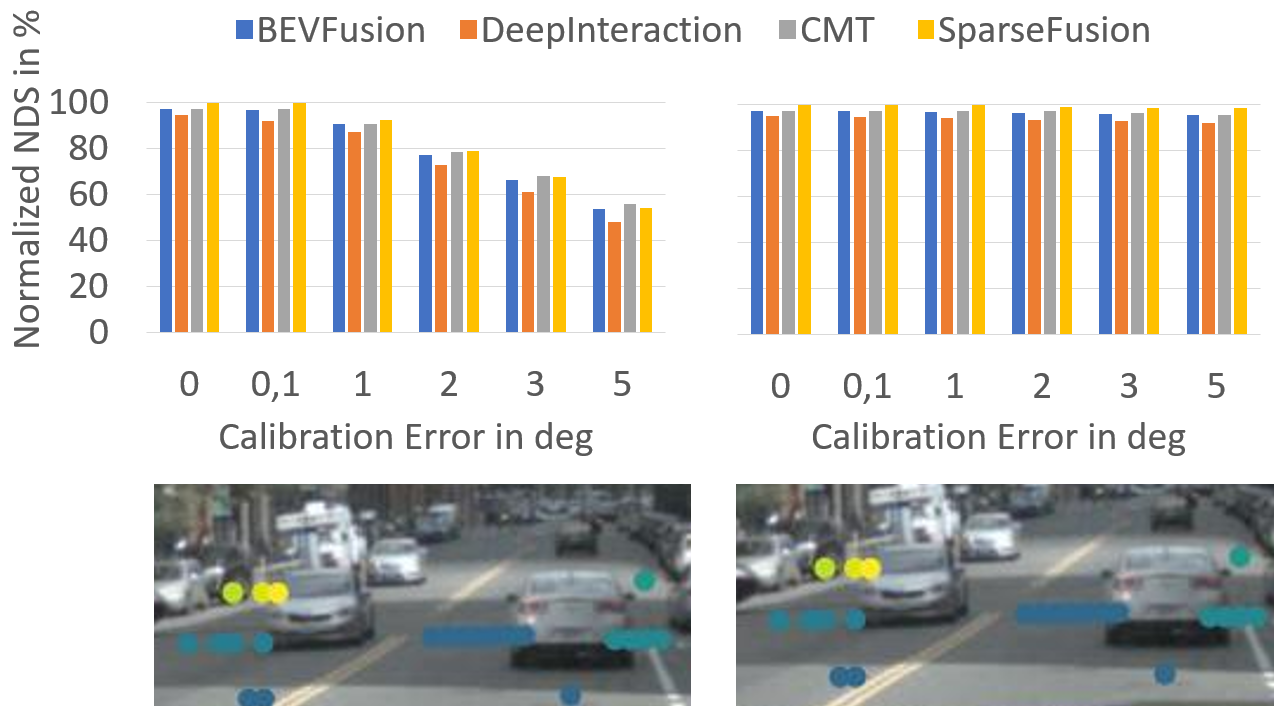}
    \caption{The plots on the left show the performance and sensor alignment in the presence of a yaw angle LiDAR-to-vehicle error. The right plots show the performance and LiDAR-camera alignment in the presence of an inverse yaw angle camera-to-vehicle error. From the LiDAR-to-camera perspective, the two sensors have identical errors. However, significant performance drops are observed only for LiDAR-to-vehicle errors.}
    \label{fig:sensivitiy}
\end{figure}

Using LiDAR as the reference sensor and enhancing detection with camera images has become the de facto standard in autonomous driving research. Fig.~\ref{fig:sensivitiy} illustrates the performance of four detection models~\cite{Xie2023, Bai2022, Liu2023, Yang2022} from the nuScenes~\cite{Caesar2019} leaderboard in the presence of various S2S calibration errors. The left image shows the models’ performance with S2S errors due to LiDAR miscalibration. The left image below shows the LiDAR-to-camera alignment in the presence of a LiDAR-to-vehicle yaw error. The upper right plot shows the models’ performance with S2S errors due to camera miscalibration. Accordingly, the lower right image shows the LiDAR-to-camera alignment in the presence of a Camera-to-vehicle yaw error. Note that both images correspond to the same yaw error of $+3$ degree for the LiDAR-to-vehicle and $-3$ degree for the Camera-to-vehicle and therefore to the exact same relative S2S error.

Even though the S2S for both configurations are the same, the primary object detection performance decline occurs in the presence of miscalibrated LiDAR, while minimal performance drops are observed in cases of camera miscalibration. This suggests that LiDAR-to-vehicle calibration plays a predominant role in overall detection performance. This finding is not unexpected, as LiDAR provides the primary features necessary for localizing objects in the environment and plays an instrumental role in evaluating object detection algorithms. This makes it clear that detecting LiDAR miscalibration is the first and foremost priority to ensure safety at all times during autonomous operation.

\subsubsection{Miscalibration Detection}
There are many robust and efficient online calibration algorithms~\cite{Schneider2017, CalibNet, RGGNet, FusionNet} available in the field of calibration research that regress the calibration parameters directly. An early consideration of calibration as a classification problem is presented by Levinson~\etal~\cite{Levinson}. The proposed method first generates an edge image and the discontinuity of the point cloud. The depth image is created by applying an edge filter,
resulting in its "edginess". The point clouds’ discontinuities are calculated by determining the depth difference between each point and its two neighbors. In~\cite{Wei},  a deep-learning-based method that can detect and calibrate misaligned sensors is introduced. Tahiraj~\etal~\cite{Tahiraj2025} also shift the focus from regressing the parameters to providing solely a classification of the miscalibration state. They propose a lightweight network based on contrastive learning to detect miscalibration in real time.
All of these approaches focus on sensor-to-sensor miscalibration detection. Yet, as previously highlighted, it is essential to move beyond detecting misalignment between sensors and toward identifying which individual sensor is miscalibrated. This work focuses on detecting angular miscalibration in LiDAR sensors.

\subsubsection{Sensor-to-Vehicle Calibration}
Several methods for calibrating individual LiDAR systems are described in the literature. Most of these approaches perform roll and pitch calibration using the ground plane, such as Multi-LiCa~\cite{Kulmer2024} and SensorX2Car~\cite{Yan2024}. Yan~\cite{Yan2024} and Seok~\cite{Seok2024} additionally provide yaw angle calibration. An alternative approach for LiDAR-to-vehicle calibration is presented in~\cite{CaLiV}. CaLiV performs LiDAR-to-vehicle calibration by leveraging calibration targets and a simple maneuver. 

Our approach to detecting miscalibration between the LiDAR and the vehicle frame differs in that we analyze the motion patterns of static objects using scene flows and train a classifier to identify miscalibration. As previously mentioned, miscalibration causes time-dependent patterns in the observed motion of objects when the vehicle is moving. Our method does not depend on known ground planes, calibration targets, or specific vehicle maneuvers.
\section{METHODOLOGY} \label{sec:methodology}
\begin{figure}[!h]
    \centering
    \includegraphics[width=1\linewidth]{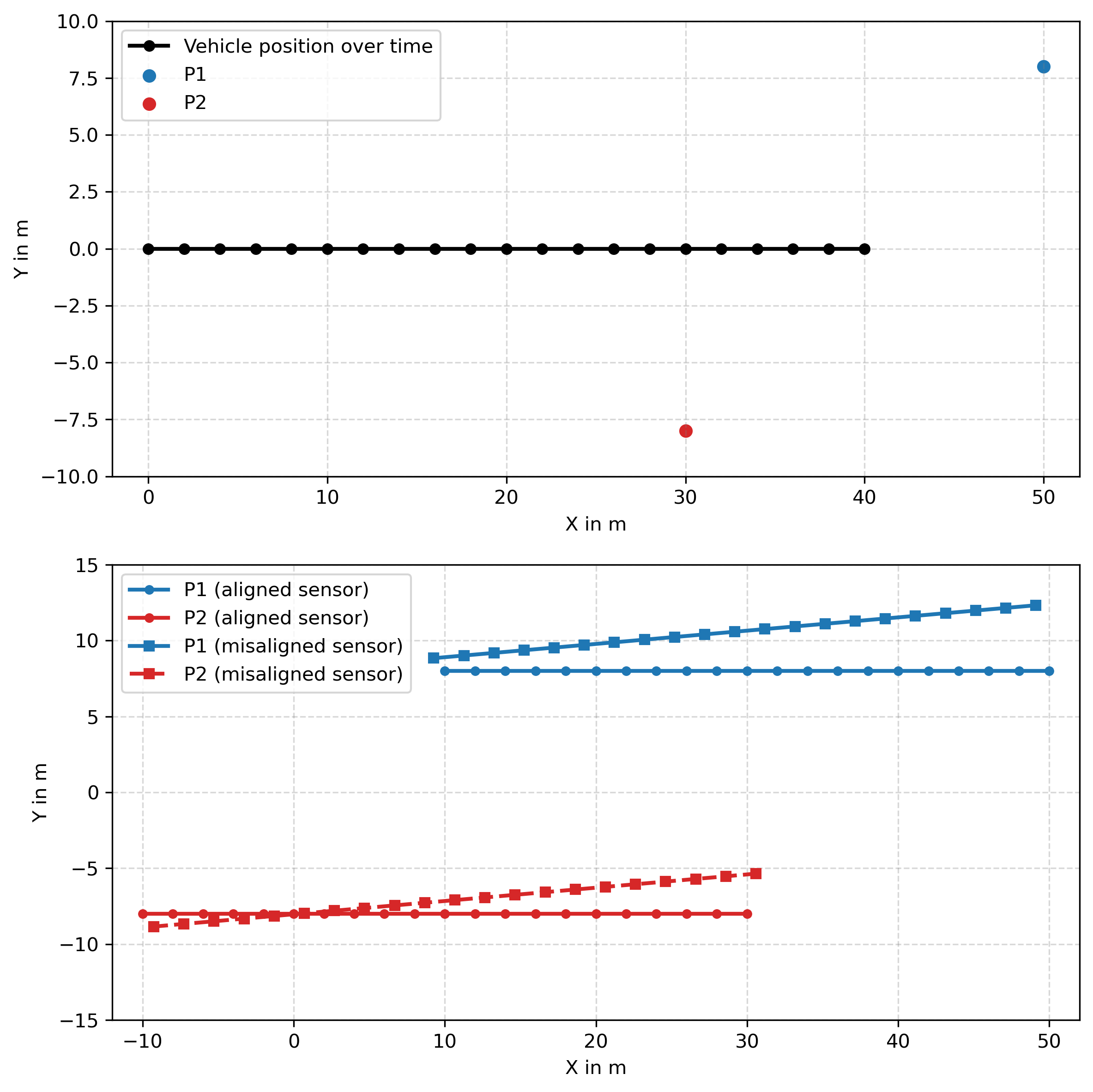}
    \caption{The idea behind using flow fields to detect LiDAR miscalibration is that a calibration error appears as a bias in the point flow. This causes distance-dependent errors that decrease as an object approaches the vehicle. The vehicle's trajectory is shown above. The resulting motion pattern of the objects, represented here as points, is shown below.}
    \label{fig:main}
\end{figure}

The underlying idea of our proposed framework is that the position of any point on a static object follows the motion pattern of the ego vehicle. If an object is directly in front of the vehicle and moving in a straight line, the trajectory of the perceived points is also expected to be straight. In the case of angular misalignment, the point's movement includes a bias in the respective dimensions and changes with the distance between the vehicle and the object. Fig.~\ref{fig:main} illustrates the variation in the perception of point positions in the case of a straight trajectory. For illustrative purposes, a straight trajectory is shown. However, it should be noted that this phenomenon applies to arbitrary trajectories.
Flow vectors incorporate information about sensor alignment status and represent the movement perceived by each sensor. The perceived flow of each point encodes ego-motion patterns that change systematically if misalignment is present.  Thus, flow provides an alternative to using hard geometric constraints~\cite{Meyer2021} or vehicle movement with SLAM~\cite{Yan2024} or ICP~\cite{Seok2024}. 

\begin{figure*}
    \centering
    \includegraphics[width=1\textwidth]{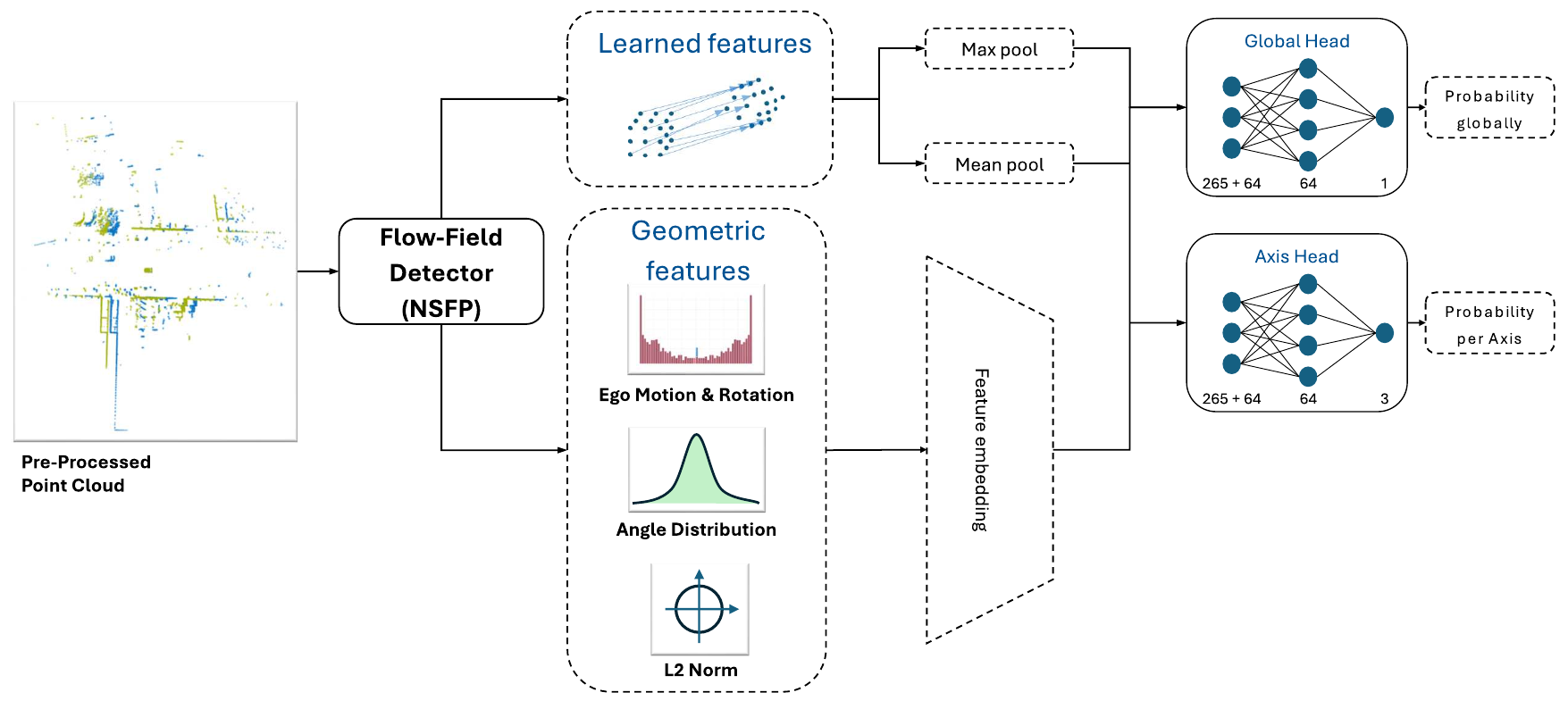}
    \caption{FlowCalib uses a two-stage learning process. First, flow fields are generated to construct features. These features are used to learn feature embeddings and train the global and axis detection heads. The heads are trained to detect miscalibration in the presence of random calibration errors and to identify the miscalibrated axis.}
    \label{fig:structure}
    \vspace{-0.5cm}
\end{figure*}

With this premise of point motion, this section describes the FlowCalib framework (see Fig.~\ref{fig:structure}) and is structured as follows: First, fault injection and the preprocessing steps that each point cloud undergoes are presented. We also explains how scene flow is generated and how the LiDAR miscalibration detection is designed. Finally, the implementation details of FlowCalib are presented.

\subsection{Fault Injection}
The nuScenes~\cite{Caesar2019} dataset is used as the base set, providing a pre-calibrated dataset. Miscalibrations are applied by injecting an additional angular error to the point cloud. The resulting transformation is given in Eq.~\ref{eq:mat_distortion}. The rotation matrix $\mathbf{R}_{\alpha_{err}}$ represents a rotation of $\alpha_{err}$ applied to the original point cloud $\textbf{X}_{org}$. The original point cloud is transformed into the distorted point cloud $\mathbf{X}_{dist}$ 

\begin{equation}\label{eq:mat_distortion}
	\mathbf{X}_{dist} = \textbf{X}_{org} \mathbf{R}_{\alpha_{err}}. \qquad
\end{equation}

Rotation is applied randomly to each axis. The resulting samples are evenly distributed across different levels of severity. $\alpha_{err} \sim \mathcal{U}\big([-5.0, -0.5]^\circ \cup [0.5, 5.0]^\circ\big)$ is sampled from a uniform distribution resulting in equal rotation exposure following~\cite{Dong2023}.
\subsection{Data Preprocessing}
The raw nuScenes point clouds require pre-processing to enable scene flow generation. The input and output of this step is illustrated in Fig.~\ref{fig:pc_overall}. The preprocessing ensures data relevance, reduces noise, and aligns the perspective with the vehicle’s reference frame. There are four key operations involved in pre-processing: 

\begin{figure}[t]
    \centering
    \begin{subfigure}{0.48\linewidth}
        \centering
        \includegraphics[width=\linewidth]{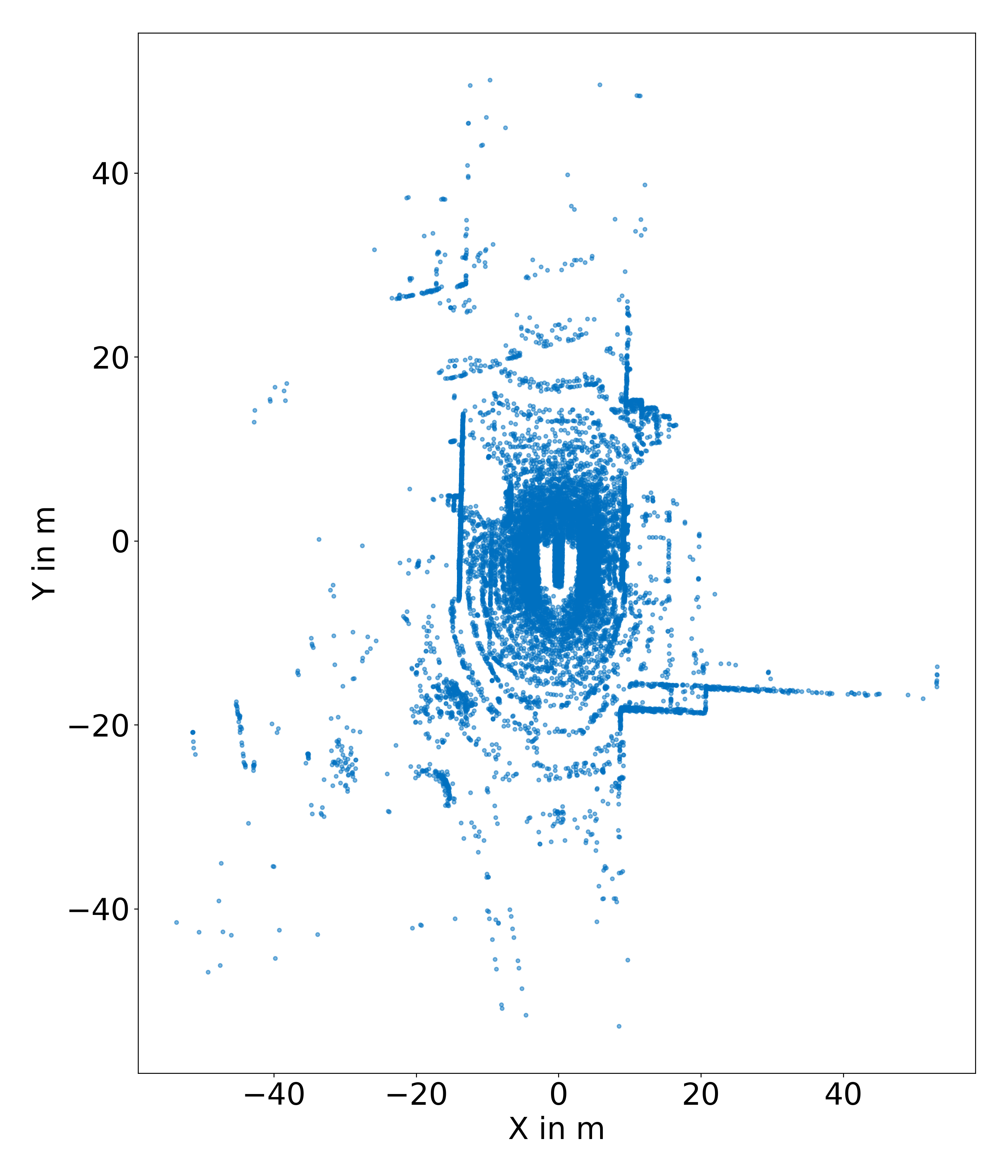}
        \caption{Before preprocessing}
        \label{fig:pc_pre}
    \end{subfigure}
    \hfill
    \begin{subfigure}{0.48\linewidth}
        \centering
        \includegraphics[width=\linewidth]{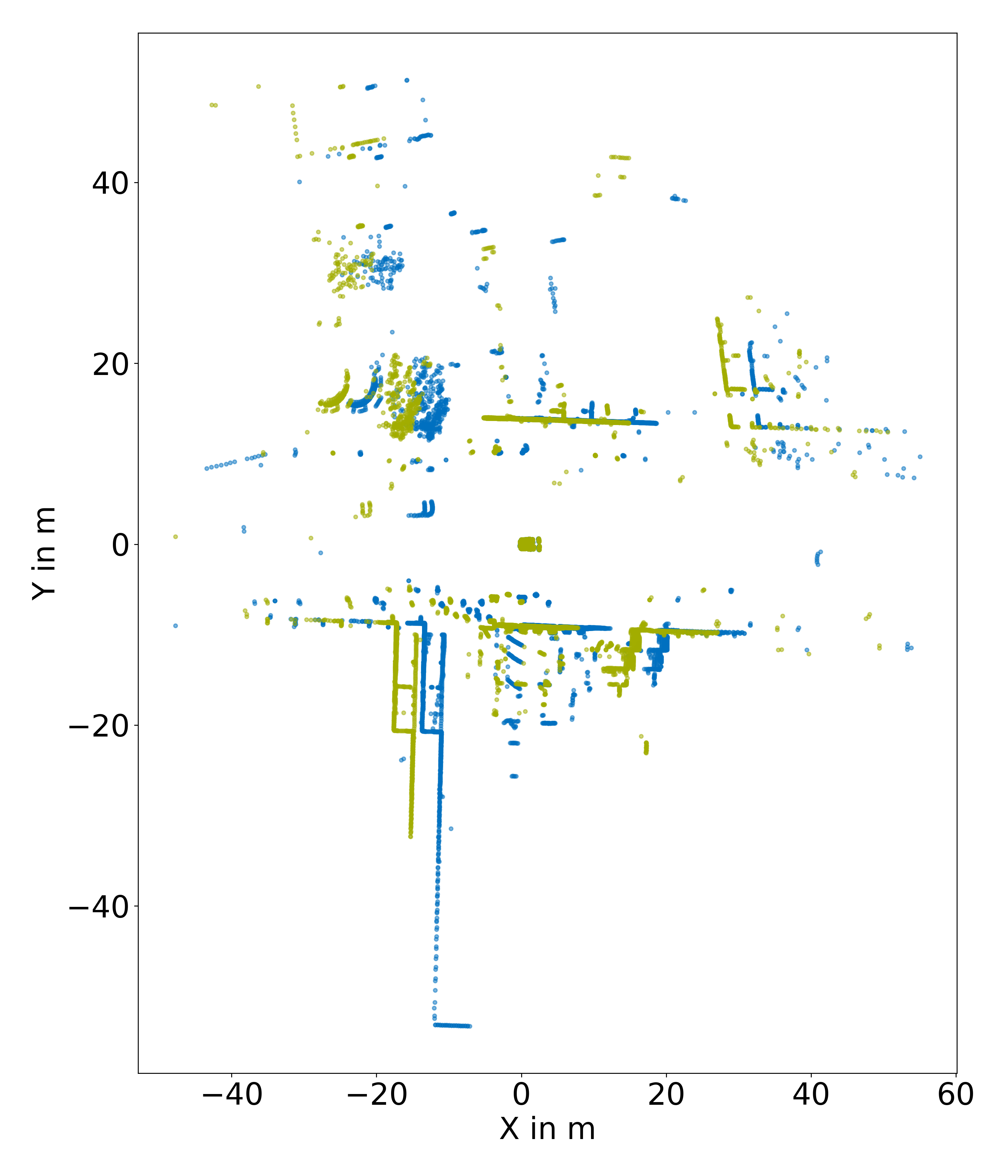}
        \caption{After preprocessing}
        \label{fig:pc_post}
    \end{subfigure}
    \caption{The point cloud is shown before (left) and after (right) the pre-processing steps. The blue and green point clouds after preprocessing are shown at timesteps $t$ and $t+1$, respectively. These serve as input for scene flow generation. Note that the point cloud are transformed into vehicle coordinate systems after the preprocessing.}
    \label{fig:pc_overall}
\end{figure}
\subsubsection{Ground removal}
Point clouds include ground points as part of the sensing process. These points represent a flat area with few distinguishing characteristics, which makes it difficult to provide robust information for scene flow prediction~\cite{Li2021}. The absence of features makes determining the movement pattern of the points difficult, as the flat surface represents a homogeneous area. Scene flow predictors have difficulty finding the correct correspondences and calculating movement over time. To overcome these challenges and prevent false flows, ground points are removed prior to scene flow generation. Ground point removal follows the implementation of UNION~\cite{Lentsch2025} employing RANSAC~\cite{Fischler1987}.

\subsubsection{Coordinate Systems}
The raw point clouds provided by the nuScenes dataset are initially represented in the LiDAR sensor frame. Howevery, the bias introduced by misalignment is anchored in the S2V transformation. To facilitate analysis and interpretation relative to the vehicle's coordinate system, the point clouds are transformed into the vehicle frame. This transformation is achieved by applying the expected extrinsic rotation, ${}^V_S\mathbf{R}$, which aligns the sensor data with the vehicle's coordinate system. This alignment enables a consistent evaluation of point positions and motions with respect to the vehicle's pose.

\subsubsection{Point Cloud Distillation}
nuScenes provides keyframes captured at 2 Hz, and additional sweeps captured at 20 Hz \cite{Caesar2019}. FlowCalib focuses on the rotation detection per keyframe. Scene flow is generated between $n_t$ time steps to incorporate temporal information described by the motion of points to detect LiDAR miscalibration. Directly estimating scene flow over large time intervals increases the risk of incorrect point correspondences, which is inherent in the scene flow generation process~\cite{Li2021}. To address this issue, scene flow is estimated over smaller steps.  Therefore, it is necessary to select frames prior to generating the scene flow. The frames chosen represent the farthest possible time steps. Key frames include ten sweeps and intermediate, non-labeled frames. Selecting $n_t$ frames results in a frequency of the chosen frames: 
\[f_{select} = \frac{f_{sampled}}{\frac{n_{frames}}{n_t}}\]
This approach enables us to generate meaningful and robust scene flow by balancing the need for temporal information with the practical challenges of point cloud matching.
\subsubsection{Dynamic Object Removal}
Moving objects can disturb the perceived movement represented in the flow field. Since they have independent velocities, point movements do not represent the sensor's perceived point flow. Therefore, to increase the robustness of miscalibration detection, it is necessary to focus on static objects. This can be achieved by removing labeled object points. In this work, these points are selected using the ground truth bounding boxes of the dataset.
\vspace{-5pt}

\subsection{Scene Flow Generation}
We use scene flow detection to generate a vector field that describes the movement of each point between two time steps. This information provides insight into the perceived movement of the LiDAR sensor. The micalibration detector. described in the next subsection, uses the vector field to check for sensor rotation because the pattern of point movement differentiates between an aligned and misaligned sensor. The scene flow generation is based on the Neural Scene Flow Prior (NSFP)~\cite{Li2021}. 

Given two consecutive 3D point clouds $\mathbf{X}_{t-1}$ and $\mathbf{X}_{t}$, NSFP aims to estimate the scene flow $F = \{\mathbf{u}_i\}_{i=1}^{\mathbf{X}_{t-1}}$, where each point $\mathbf{x}_{t-1} \in \mathbf{X}_{t-1}$ moves according to a flow vector $\mathbf{u} \in \mathbb{R}^3$ such that $\mathbf{x}'_{t-1} = \mathbf{X}_{t-1} + \mathbf{u}$. The problem is formalized as minimizing the distance between the transformed points from $\mathbf{X}_{t-1}$ and the points in $\mathbf{X}_{t}$, with an added regularization term:
\begin{equation}
\mathcal{F}^* = \arg\min_\mathcal{F} \sum_{\mathbf{x}_{t-1} \in \mathbf{X}_{t-1}} D(\mathbf{x}_{t-1} + \mathbf{u}, \mathbf{X}_{t}) + \lambda C,
\end{equation}
where:
\begin{itemize}
    \item $D(\mathbf{x}_{t-1} + \mathbf{u}, \mathbf{X}_{t})$ computes the distance from the shifted point $\mathbf{x}_{t-1} + \mathbf{u}$ to its nearest neighbor in $\mathbf{X}_{t}$,
    \item $C$ is a regularization term,
    \item $\lambda$ is a weighting factor for the regularizer.
\end{itemize}
For the distance function, the squared Euclidean distance to the nearest neighbor is used. NSFP additionally incorporates a neural network as an implicit regularizer that optimizes the scene flow. This allows scene-specific flows at runtime and does not require pre-training or labeled datasets~\cite{Li2021}.

\subsection{Miscalibration Detection}
A prediction model using a mixture of hand-crafted and learned feature embeddings is proposed to check for miscalibration of the LiDAR. First, the flow field $\mathcal{F}$ is generated by the scene flow distillation process using the point cloud at time step $t_n$, which is used to anchor the flow vectors in the correct spatial context. Each flow vector ${\mathbf{u}_{t-1 \to t, i}} \in \mathcal{F}_{t-1 \to t}$ corresponds to a point $\mathbf{x}_i$ in the point cloud $\mathbf{X}_{t-i}$, representing the motion of $\mathbf{x}_i$ from time $t-1$ to $t$.

The detector consists of two encoders that learn global flow feature embeddings and handcrafted geometric flow features, as well as two separate decoders that detect the LiDAR's alignment. The detector's output is an alignment indicator in the range of $[0, 1]$, representing the degree of trust in the angular alignment of each axis.

\subsubsection{Global Flow Features}
We employ a PointNet architecture~\cite{Charles2017} to process unordered flow vectors (see Fig.~\ref{fig:flow_main}) and extract global flow features. The feature dimension is expanded by processing each flow vector through a series of  $1 \times 1$ convolutions. Each convolutional layer is followed by batch normalization and a ReLU activation to help stabilize and accelerate training. The output represents a global signature of the flow field illustrated as the Learned features box in Fig.~\ref{fig:structure}. To aggregate these features, maximum and mean pooling are applied. Maximum pooling captures dominant patterns, but it is sensitive to outliers. Mean pooling, on the other hand, provides a smoother, more robust representation. Combining the two mitigates the impact of local inaccuracies in the flow field, particularly in regions with dense objects where backward flows can introduce ambiguities. Unlike the original PointNet, we omit concatenating local and global features. This design choice reduces the number of model parameters and computational complexity because the model focuses on the global characteristics of the flow field for misalignment detection.

\subsubsection{Geometric Features}
In addition to learned global features, specific geometric features that are directly related to the sensor's misalignment state are extracted. Computing the global statistics of these features captures the overall effect of misalignment on the point cloud while mitigating the impact of outliers. Each feature is produced for every axis individually and then concatenated.
\begin{figure} [h!]
    \centering
    \begin{subfigure}{\linewidth}
        \centering
        \includegraphics[width=1.0\linewidth]{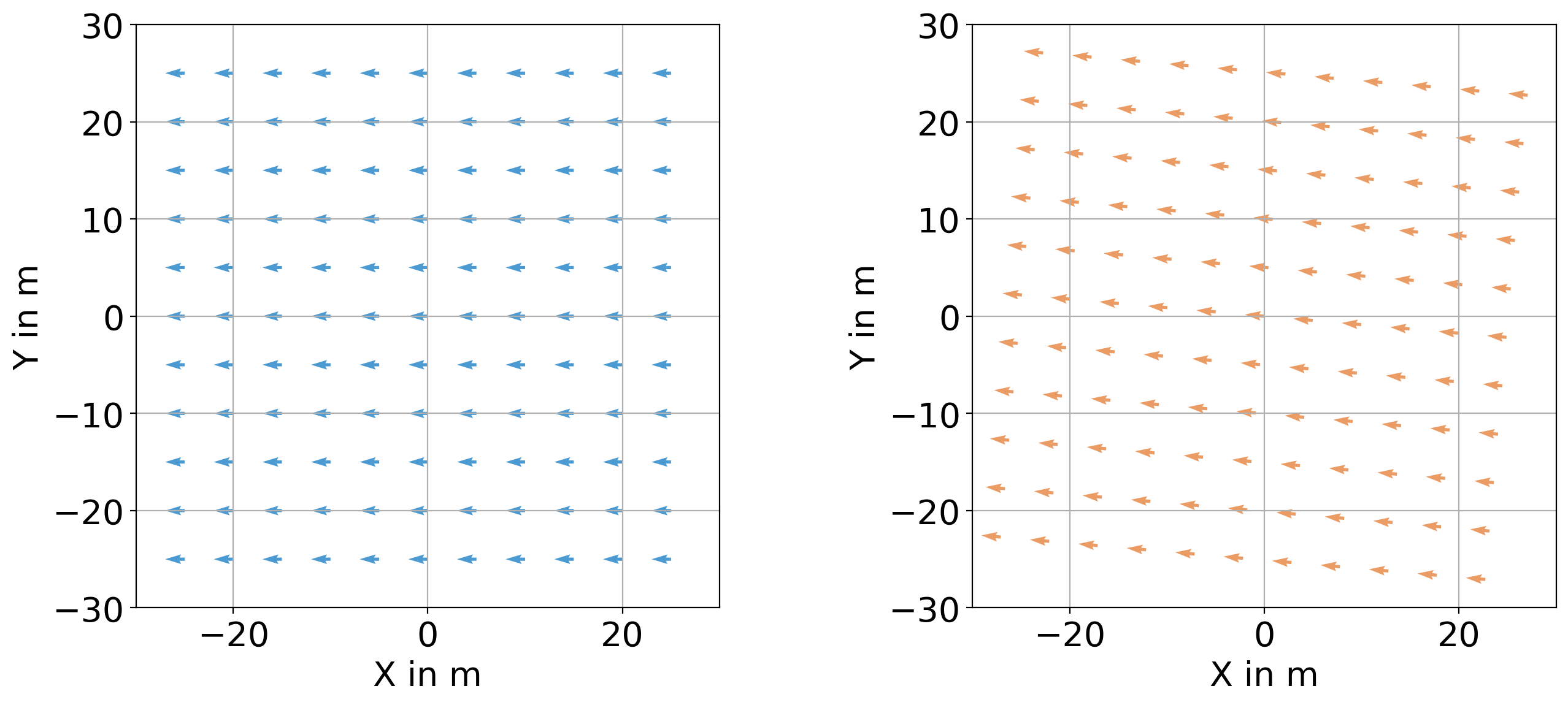}
        \label{fig:traj}
    \end{subfigure}

    \begin{subfigure}{\linewidth}
        \centering
        \includegraphics[width=1.0\linewidth]{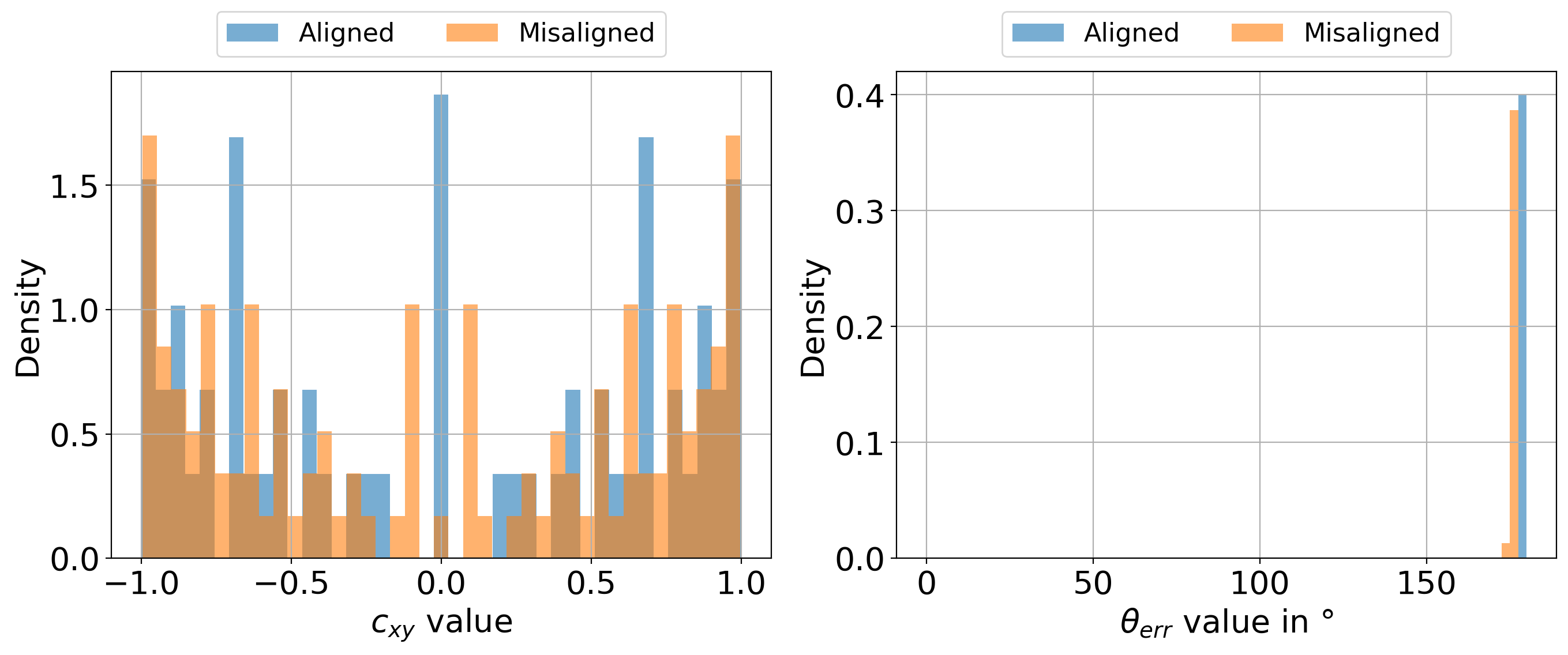}
        \label{fig:perc_points}
    \end{subfigure}

    \caption{\textbf{Above}: This illustrates the flow fields of aligned and misaligned sensors. \textbf{Below}: The resulting distribution of the angle $\psi$ and cross value $c$ for aligned and misaligned flow fields.}
    \label{fig:flow_main}
    \vspace{-5pt}
\end{figure}
\textbf{Magnitude.} The flow magnitude for each point is computed as the Euclidean norm of its flow vector $m_i = \left \| \mathbf{u}_i \right \| \in \mathbb{R}$. The resulting vector $ \mathbf{m} \in \mathbb{R}^{N}$ captures the overall motion length of each point in $\mathbf{X}_i$ and implicitly reflects the expected impact of misalignment. 
Sensor misalignment introduces an additional lateral offset in the flow field, which is proportional to the forward ego-motion. Therefore, the greater the ego-vehicle's forward movement, the more pronounced the effect of misalignment on the flow magnitude. To capture the global effect, the mean $\mu_{\mathbf{m}}$ and standard deviation $\sigma_{\mathbf{m}}$ of $\mathbf{m}$ are taken into consideration.

\textbf{Angle.}
The angle between each flow vector's $y$$z$, $x$$z$, and $x$$y$ axes provides a direct measure of the sensor's offset relative to the vehicle's forward movement. For each flow vector the angle $\psi_i = \arctan{\frac{u_{d_n, i}}{u_{d_m, i}}}$ is calculated. With $d$ representing the respective axis for pitch, roll, and yaw.

The angles are aggregated into a histogram $\mathbf{h}_{n_{bins}} \in \mathbb{R}^{{n_{bins}}}$, providing a comprehensive representation of the angle distribution (Fig.~\ref{fig:flow_main}). Therefore, histogram-based angular features are used. This enables the model to learn the characteristic bias introduced by the misalignment angle $\alpha_{err}$. The number of bins $n_{bins}$ is a hyperparameter that controls the histogram's resolution. 

\textbf{Rotation.}
To capture rotational effects, the cross product between each point's radial position relative to the ego-vehicle ${}^{V_i} {P(t_0)_{d_nd_m,i}}$ and its flow vector ${}^{V_i} {\mathbf{u}_{t \to t+1, i}}$ is computed as defined in Eq.~\ref{eq:point_flow_cross}. The distribution of $\mathbf{c_{d_n d_m}} \in \mathbb{R}^{N}$ values provides a global descriptor of the detected movement. Since misalignment introduces a systematic offset in the flow field, the model uses the mean and standard deviation to learn the effects of the induced bias. This is important as the sensor misalignment introduces a global offset to the flow field, which is captured by the distribution of ${\mathbf{c}_{d_n d_m}}$. 

\begin{equation}\label{eq:point_flow_cross}
\begin{split}
	{c_{d_n d_m,i}} 
	&= {}^{V_i} {\mathbf{p}(t_0)_{d_n d_m,i}} \times {}^{V_i} {\mathbf{u}_{t_0 \to t_1,i}} \\
	&= |{}^{V_i} {\mathbf{p}(t_0)_{d_n d_m,i}}| \, |{}^{V_i} {\mathbf{u}_{t_0 \to t_1,i}}| \sin(\phi)
\end{split}
\end{equation}
The distribution is illustrated in Fig~\ref{fig:flow_main}. The resulting geometrical features are concatenated, representing a per-sample $21 + 3*n_{bins}$ dimensional feature vector $\mathbf{f}_{geom} \in \mathbb{R}^{21 + 3*n_{bins}}$. This vector is fed through a three-block MLP, each block consisting of a linear layer, batch normalization, and ReLU activation. The dimensions of the linear layers are 256, 128, and 128. The result is the embedded feature vector $\mathbf{f}_{emb, geom} \in \mathbb{R}^{128}$. 

The global flow and geometric features are concatenated and passed through additional MLPs. Two detection heads are implemented, each consisting of a linear layer, an activation function, and an additional linear layer. The head responsible for miscalibration detection has an output dimension of three, which allows it to relate the prediction to pitch, roll, and yaw. A value close to 1 indicates high confidence in misalignment, and a value close to 0 indicates alignment.  
\subsection{Implementation Details}
FlowCalib is implemented in pytorch~\cite{Paszke2019}. ReLU is used as the activation function. The loss is represented by a Binary Cross Entropy (BCE) with logits. This implementation allows for more stable and integrated usage of ReLU and BCE than separate usage.
AdamW~\cite{Loshchilov2019} is used as the optimizer with a learning rate of $8e-3$. The optimizer uses a weight decay of $1 \times 10^{-4}$. To maintain a low dimension of the geometric encoders while providing a detailed picture, $n_{bins}$ is set to 72, leading to a resolution of $ 5^\circ$.

Global features are encoded by a two-block encoder that consisting of a $1 \times 1$ convolution, a 1D batch normalization, and an activation function. The channel sizes used are 64 and 128. The geometric feature encoder uses a linear layer instead of convolutions. The layers have depths of 256, 128, and 128. The decoder is a block consisting of a linear layer, an activation function, and an additional linear layer that reduces the output to 1 or 3, respectively. 
\section{RESULTS \& DISCUSSION}
This section presents the performance of global misalignment detection, as well as the model's ability to verify proper sensor alignment. It also presents the detection performance on the individual axes. 
\subsection{Global Alignment Status}
Table~\ref{table:per_angle_all} reports the detection capabilities, categorized based on the severity of distortion. The models achieve an overall accuracy of 81.16 \%. Angular miscalibrations in the range of $[\pm5, \pm2]^\circ$ can be detected with high accuracy of 90.27\%. The classification performance drops to 73.81\% for angular miscalibrations in the range of $[\pm2, \pm0.5]^\circ$, showing comparable accuracy in the Medium and Hard distortion severity levels. 
 
 \begin{table}[h]
 	\centering
    \begin{tabular}{l l r r c}
    \toprule
        \textbf{Category} & \textbf{Rot. Error} & \textbf{Correct} & \textbf{Incorrect} & \textbf{Correct in \%} \\
	 	\midrule
	 	Aligned     & $[-0.4, 0.4]^\circ$ & 527 & 187 &  73.81 \\
	 	\midrule
	 	\quad Hard  & $(\pm1, \pm0.5]^\circ$&  173 &  44 &  79.72 \\
	 	\quad Medium & $(\pm2, \pm1]^\circ$ &  87 &   22 &  79.82 \\
	 	\quad Easy  & $[\pm5, \pm2]^\circ $ &  566 &  61 &  90.27 \\
	 	\midrule
	 	Total & & 							   1353 & 314 &  81.16 \\
	 	\bottomrule
	 \end{tabular}
	 \caption{Analysis of the global classification performance on different levels of severity. Note that hard refers to errors with only slight angular velocities that are harder to detect.}
	 \label{table:per_angle_all}
\end{table}
To better understand the performance drop, Table~\ref{tab:axis_rotation_results} reports the model’s true-positive detection performance for global misalignment at the specific rotation axes where the error is introduced. 
In other words, the table shows for which samples the model struggles to detect a miscalibration.

Table~\ref{tab:axis_rotation_results} shows the number of samples in the evaluation dataset for specific combinations of angular miscalibration. The analysis concludes that the model has difficulty classifying misalignments arising from pitch angle rotations. Only 24.73\% of samples are classified correctly when miscalibration occurs in that dimension. The share of correctly classified samples for combinations in which another angular miscalibration is present becomes much higher. When all three angles are disturbed, a high detection rate of 94.70\% can be achieved. One possible reason is the strong overall distortion caused by misalignment in three dimensions.

\begin{table}[htbp]
\centering
\begin{tabular}{l c c c}
\toprule
\textbf{Category} & \multicolumn{1}{c}{\textbf{\begin{tabular}{c}Correct \\ Samples\end{tabular}}} 
                  & \multicolumn{1}{c}{\textbf{\begin{tabular}{c}Incorrect \\ Samples\end{tabular}}} 
                  & \multicolumn{1}{c}{\textbf{\begin{tabular}{c}Share correct \\ in \%\end{tabular}}} \\
\midrule
$\phi$                  & 104 &   8 & 92.86 \\ 
$\theta$                &  23 &  70 & 24.73 \\
$\psi$                  & 112 &   5 & 95.73 \\
$\phi$  \& $\theta$     &  92 &   8 & 92.00 \\
$\phi$ \& $\psi$        & 102 &   7 & 93.58 \\
$\theta$ \& $\psi$      &  89 &  12 & 88.12 \\
$\phi$ \& $\theta$ \& $\psi$ & 304 &  17 & 94.70 \\
\bottomrule
\end{tabular}
\caption{Classification results in the presence of different $\phi$ (Roll), $\theta$ (Pitch), $\psi$ (Yaw) error combinations.}
\label{tab:axis_rotation_results}
\end{table}

Although the model does not perform well in every dimension, its ability to detect misalignment is evident. The model has difficulty predicting deviations in the pitch angle in particular. There are several possible reasons for this behavior. Distortions in the pitch angle lead to uniform up- or down-movements. Yaw angles vary more distinctively relative to the vehicle depending on the position of the point. For example, a point on the front of the vehicle is shifted left or right, while a point on the side of the vehicle moves forward or backward. Similarly, the roll error effect depends on the position of the point. These variations create distinctive patterns that the model can detect. However, as this is not the case for pitch perturbations, it becomes more difficult for the model to detect them.

\subsection{Rotated Axis detection}
In the following, the model's ability to detect the distorted axis, as represented by the output of the axis head, is analyzed. 

As shown in Table~\ref{tab:axis_metrics}, across all data samples, the model correctly detects present rotations around the pitch axis in 60.81\% of the cases. Pitch misalignment is more difficult to detect because it produces a very uniform motion pattern across the entire scene.

In contrast, the model performs better at identifying present rotations around the roll and yaw axes, achieving accuracies of 87.04 \% and 76.06 \%, respectively. 

\begin{table}[htbp]
	\centering
	\begin{tabular}{lcccc}
		\toprule
		Axis  & Accuracy & Precision & Recall \\
		\midrule
		$\phi$  & 87.04 & 86.60 & 78.50  \\
		$\theta$ & 60.81 & 55.87 & 73.50  \\
		$\psi$   & 76.06 & 63.82 & 88.73  \\
		\bottomrule
	\end{tabular}
	\caption{Axis-specific detection results in the presence of different $\phi$ (Roll), $\theta$ (Pitch), $\psi$ (Yaw) error combinations.}
	\label{tab:axis_metrics}
\end{table}

The model reliably detects misalignment and performs particularly well when distortions span multiple angles. In such instances, larger deviations across several dimensions generate clearer misalignment patterns in the flow field, which the model effectively leverages. The model also shows promise in identifying the axis of decalibration.

\section{CONCLUSION}
In this work, we introduced FlowCalib, the first framework for detecting LiDAR-to-vehicle miscalibration by leveraging scene flow patterns of static objects.  FlowCalib provides a novel and lightweight alternative to existing calibration or miscalibration-detection approaches. We exploit point-wise motion cues without relying on additional sensors or strict geometric priors. Our results on the nuScenes dataset demonstrate that flow-based cues contain meaningful signatures of angular misalignment and that these can be learned to reliably identify global and axis-specific calibration errors. In particular, FlowCalib achieves strong performance in detecting yaw and roll miscalibration, with consistently high detection rates when multiple axes are jointly perturbed. These findings confirm that motion-induced distortions in the flow field provide a robust and interpretable basis for identifying extrinsic sensor calibration errors during autonomous vehicle operation.

%%%%%%%%%%%%%%%%%%%%%%%%%%%%%%%%%%%%%%%%%%%%%%%%%%%%%%%%%%%%%%%%%%%%%%%%%%%%%%%%

\bibliography{FlowCalib.bib}
\bibliographystyle{IEEEtran}

\end{document}